\documentclass{article}
\usepackage{hyperref}
\usepackage{xspace}
\usepackage{amsmath}
\usepackage{edsstuff}
\usepackage[square,comma,sort&compress,numbers]{natbib}

\newtheorem{algorithm}{Algorithm}
\def\comment#1{\vspace{-2ex}\par\noindent{\bf Comment:} #1}
\def\Alg#1{Algorithm \ref{alg:#1}}

\begin{document}

\title{Improved RANSAC performance using simple, iterative minimal-set solvers}

\author{
E. Rosten\footnote{\texttt{er258 at cam.ac.uk}, Department of Engineering, University of Cambridge, UK}, 
G. Reitmayr\thanks{\texttt{reitmayr at icg.tugraz.at} Technische Unversitaet Graz, Graz, Austria},
T. Drummond\footnote{\texttt{twd20 at cam.ac.uk}, Department of Engineering, University of Cambridge, UK}
}

\maketitle

\begin{abstract}
RANSAC is a popular technique for estimating model parameters in the presence of
outliers. The best speed is achieved when the minimum possible number of points
is used to estimate hypotheses for the model.  Many useful problems can be
represented using polyomial constraints (for instance, the determinant of a
fundamental matrix must be zero) and so have a number of soultions which are
consistent with a minimal set.  A considerable amount of effort has been
expended on finding the constraints of such problems, and these often require
the solution of systems of polynomial equations.  We show that better
performance can be achieved by using a simple optimization based approach on
minimal sets.  For a given minimal set, the optimization approach is not
guaranteed to converge to the correct solution. However, when used within RANSAC
the greater speed and numerical stability results in better performance overall,
and much simpler algorithms.  We also show that by selecting {\em more} than the
minimal number of points and using robust optimization can yield better results
for very noisy by reducing the number of trials required.  The increased speed
of our method demonstrated with experiments on essential matrix estimation.

\end{abstract}

\section{Introduction}

\def\Qt{\ensuremath{\tilde{Q}}\xspace}
\def\Et{\ensuremath{\tilde{E}}\xspace}
\def\Bz{\mbox{$\pmb 0$}}

Many computer vision systems operating on video require frame-rate operation in
order to be useful. This paper is concerned with estimating parameters (in
particular, the essential matrix) with the greatest possible efficiency.
For RANSAC~\cite{ransac} schemes to be efficient, it is important to be
able to estimate model hypotheses using the smallest possible amount of data,
since the probability of selecting a set of datapoints without outliers
decreases exponentially with the amount of data required. A collection of
datapoints of the minimum required size is known as a {\em minimal set}.  In
some computer vision problems, (such as essential matrix
estimation~\cite{nister5point} and image stitching with radial
distortion~\cite{byrod09minimal}), the data describe a system which is subject to a number of
polynomial constrains.  Therefore, direct minimal set algorithms involve finding the
solution of polynomial sets of equations.

This paper is about using iterative solvers instead of direct polynomial solvers and
is motivated by the following observations:
\begin{itemize}
\item Minimal set algorithms are useful when one is performing robust
estimation using RANSAC or a related scheme.

\item In RANSAC, {\bf speed matters even at the expense of quality}. If one can
conceive of an optimization which quadruples the number of hypotheses which can
be generated and tested within a given time budget, even if three fifths of the
hypotheses are bad, there is still a net increase in performance. 

\item Finding the roots of high-degree polynomials is notoriously
hard~\cite{nr_in_C_root} as the numerical stability of roots is very poor.
Therefore, even direct solvers will not necessarily converge to correct
solutions.

\item There is no escape from iterative algorithms, as there are no general
closed-form solutions for polynmials of degree five and higher.

\item If one picks a super-minimal set of points, the
probability of having at least a minimal number of inliers is much higher.

\item There are many problems for which no known direct minimal algorithms
exist.

\end{itemize}
Therefore, we propose two approaches: 
\begin{enumerate}
\item Pick a minimal set and the model using
a simple, unconstrained nonlinear optimizer. See \Alg{lm} and \ref{alg:nuts}
in \Sec{nonlinear}.
\end{enumerate}
There are a number of theoretical trade-offs beteween optimization and
polynomial based approaches. Both methods may not yield the correct answer even
with a minimal set of inliers.  

Optimization is numerically stable, but gives at most one answer, whereas
polynomial methods will not converge successfully if the correct root is poorly
conditioned. In some important cases (\eg essential matrix estimation)
optimization has three advantages: the algorithm is simpler, faster and more
numerically stable.

\begin{enumerate}
\stepcounter{enumi}
\item 
Pick a super-minimal set and estimate the model 
using a robust algorithm such as iterative reweighted least squares. 
See \Alg{robust} in \Sec{nonlinear}.
\end{enumerate}
In the presence of high outlier levels, the probability of having at least
enough good points within the super-minimal set is much higher than the
probability of picking a minimal set of only good points. This difference becomes
high enough to outweigh the slow speed and poor convergence of robust
optimization.  An analogy can be drawn to forward error correction: by
using a redundant representation (the super-minimal set), errors can be
tolerated and it does not matter where the errors occur.

In this paper, we apply these methods to the
estimation of essential matrices.
Essential matrices are often found from as set of correspondences
between points in two images of the same scene. 
They are to estimate efficiently
because the
data from point correspondences contains outliers, and the minimal set is quite
large (5 points).

Robust estimation methods such as
M-estimation~\cite{huber64estimation,huber81statistics,torr97development}, case
deletion and explicitly fitting and removing outliers~\cite{sim06infinity}, can
be used but these often only work if there are relatively few outliers.  So the
essential matrix is often found using some variant of
RANSAC~\cite{ransac,torr97development} (RANdom SAmple Consensus) followed by an
iterative procedure such as M-estimation in order to robustly minimize the
reprojection error using all the data. 

The
essential matrix has five degrees of freedom and the minimal set is five point
matches.  The five matches yield up to 10 solutions (see \eg
\cite{heyden99reconstruction} for a recent proof). A number of practical
algorithms have been proposed~\cite{triggs20routines,philip96algorithm}, the
most prominent of which (due to its efficiency) is the `5-point algorithm',
proposed by Nist\'er~\cite{nister5point}. 

The 5-point algorithm involves setting up and solving a system of polynomial
equations, so a number of  related alternatives have been proposed which
generally attempt to simplify or sidestep that process. 

A number of related alternatives have been proposed, which trade
speed for simplicity.
For instance,
Gr\"obner bases can be used to solve the polynomial equation system~\cite{stewenius06fivepoint}
(requiring a $10\times10$ eigen decomposition), as can the
hidden variable resultant method~\cite{li06five} or 
a nonlinear eigenvalue
solver~\cite{kukelova08polynomial}.  The problem of solving sets of polynomial 
equations can be sidestepped by reformulating the problem as a constrained
function optimization~\cite{batra07alternative}. 

Some approaches to getting faster performance make use of constrained motion
~\cite{scaramuzza09monocular,ortin01indoor} in order
to 
reduce the size of the minimal set required. These are 
therefore not applicable to general use.

%
%

We compare our algorithms to the 5-point algorithm (\Alg{nister}).  Since speed
is critical in determining performance, we describe our implementations of the
nonlinear optimization and 5-point algorithms in Sections \ref{sec:nonlinear}
and \ref{sec:nister5} respectively, in addition to providing the complete
source code as supplemental material. Results are given in \Sec{results}.

\section{Optimization based solvers}
\def\trans{\ensuremath\hat{t}}
To optimize an essential matrix, we use a minimal  (\ie 5 degree of freedom), unconstrained
parameterization related to the one presented in \cite{skarbek09epipolar}.
An essential matrix can be constructed of a translation and a rotation:
\begin{equation}
	E = [\Vec{\trans}]_\times R,
\end{equation}
where $\Vec{\trans}$ is a unit vector,  $R$ is a rotation matrix
and $[\Vec{\trans}]_\times$ is a matrix such that for any vector $\Vec{v}\in {\mathbb R}^3$,
$[\Vec{\trans}]_\times \Vec{v} = \Vec{\trans}\times \Vec{v}$. Given two 2D views of 
a point in 3D, as the homogeneous vectors $\Vec{p}$ and $\Vec{p}'$, the residual error with respect to an estimated essential
matrix $\hat{E}$ is given by:
\begin{equation}
	r = \Vec{q}'{}\Trn \hat{E} \Vec{q}\label{eqn:residual}.
\end{equation}
We represent $R$ with the 3-dimensional Lie group, SO(3) (see
\eg~\cite{sattinger1986lie,varadarajan1974lie}). With the exponential map
parameterization, we choose the three generators to be:
\[
\newcommand\SmallMatrix[1]{\left[\begin{smallmatrix}#1\end{smallmatrix}\right]}
G_1 = \SmallMatrix{
      0  &  1 &  0\\
	  -1 &  0 &  0\\
	  0  &  0 &  0
      },\ \ 
G_2 = \SmallMatrix{
      0  &  0  &  1\\
	  0  &  0  &  0\\
	  -1 &  0  &  0
      },\ \ 
G_3 = \SmallMatrix{
      0  &  0  &  0\\
	  0  &  0  &  1\\
	  0  & -1  &  0
      }.
\]
By taking infinitesimal motions to be left multiplied into $R$,
the
three derivatives of $r$ with respect to $R$ are:
\begin{equation}
\Vec{q}' [\Vec{\trans}]_\times G_i R \Vec{q} \label{eqn:paramr}, \qquad i \in \{1, 2, 3\}.
\end{equation}
We parameterize
$\Vec{\trans}$ using a rotation so that $\Vec{\trans} = R_t [1\ 0\ 0]\Trn$,
with infinitesimal motions right multiplied into $R_t$.
The remaining \emph{two}
derivatives are therefore given by:
\begin{equation}
\def\colvec{\left[\raisebox{-1.5ex}{$\stackrel{1}{\stackrel{0}{\scriptstyle 0}}$}\right]}
\Vec{q}'\left[R_t G_i \colvec \right]_\times R\Vec{q}\label{eqn:paramt}, \qquad i \in \{1, 2\},
\end{equation}
since $[1\ 0\ 0]\Trn$ is in the right null
space of $G_3$. Note that the resulting optimization
does not need to be constrained. The epipolar reprojection errors (the distance between a
point and the corresponding epipolar line, not the `gold-standard' reprojection error), $g$ are given by:
\begin{equation}
	\def\proj{\left[\rule{0ex}{2.15ex}\raisebox{-.5ex}{$\smash{\stackrel{1\ 0\ 0}{\scriptstyle 0\ 1\ 0}}$}\right]}
	g  =  \frac{r}{\abs{\proj E\Vec{q}}},\qquad g' = \frac{r}{\abs{\proj E\Trn \Vec{q}'}}.
	\label{eqn:gold}
\end{equation}

\begin{algorithm}
Pick a minimal set of points, a random translation direction, a random rotation
and minimize the sum of squared residual errors (\Eqn{residual}) using the LM (Levenberg-Marquadt)
algorithm. Hypotheses  that fail to converge quickly or converge with a large
residual error are rejected.  \label{alg:lm}
\end{algorithm}

\comment{This is the standard algorithm for solving least-squares problems. 
In practise, the method is very insensitive to the choice of inital rotation.
This
technique yields zero or one solutions.}

\begin{algorithm}
Pick a minimal set of points and minimize the sum of squared residual errors
using Gauss-Newton, abandoning hypotheses which do not converge sufficiently
quickly.
\label{alg:nuts}
\end{algorithm}
\comment{ Although Gauss-Newton does not converge as effectively and reliably
as LM, the low overhead means that the algorithm can converge in much less
time.  Additionally failure can be very fast, so little time is wasted on cases
where the optimization may be very slow.  {\bfseries\textsl{In our tests with real data,
this is the best performing algorithm.}}
}

\begin{algorithm}
Pick a non minimal set and minimize the reweighted sum-square epipolar
reprojection error (\Eqn{gold}) using LM. Hypotheses with a large residual
error are rejected.
\label{alg:robust}
\end{algorithm}
\comment{A non-minimal set has a higher probability of containing at least five good
matches compared to a minimal set. Since reweighted least-squares is robust to
outliers~\cite{nr_in_C_opt}, the algorithm can converge to the correct answer even in the presence
of errors.}

\label{sec:nonlinear}

\section{Our implementation of the 5-point algorithm}

\newcommand\hilight[1]{{\sl #1}}

\label{sec:nister5}
Our implementation follows the implementation in \cite{nister5point}, with some
differences which improve the speed/accuracy trade-off.
Below is an outline of the algorithm, with our
modifications highlighted in oblique text.
Given two views of a point in 3D, $\Vec{q}$ and $\Vec{q}'$, the points are related with the
essential matrix $E$:
\begin{equation}
	\Vec{q}' E \Vec{q} = 0\label{eqn:ess}
\end{equation}
By defining $\Et = [ E_{11},  E_{12}, E_{13}, E_{21}, \cdots]\Trn$ 
and \mbox{$\tilde{q} = [ q'_1q_1,\ q'_1q_2,\ q'_1q_3,\ q'_2q_1,\cdots]$},
\Eqn{ess} can be
rewritten as the vector equation $\tilde{\Vec{q}} \Et = 0$. Stacking five sets of
equations from five different points gives the homogeneous set of equations $
	\Qt \Et = \Bz$,
where \Qt is a $5\times9$ matrix.
The elements of $E$, \Et lie in the four
dimensional null space of \Qt. If we were to extract the null space using
singular value decomposition, it would be the single most expensive part of the
algorithm.
\hilight{
Since later stages of the algorithm do not
require an orthonormal basis for the null space, we have found that best performance
is achieved by using Gauss-Jordan reduction.}
Using elementary row operations, \Qt is reduced to $[I | A ]$. Since:
\begin{equation}
	[I | A] \left[\frac{\displaystyle A}{\displaystyle -I}\right] =\Bz,
\end{equation}
the matrix $[A\Trn|-I ]$ spans the null space of \Qt. The computational cost is
that of computing a $5\times5$ matrix inverse. Since \Et can be written as a linear
combination of the four vectors spanning the null space:
\begin{equation}
	\Et\Trn = [x\ y\ z\ 1]\left[A\Trn|-I\right]\label{eqn:exyz},
\end{equation}
what remains is to find $x$, $y$ and $z$.

There are 10 cubic constraints on an essential matrix given by
$|E| = 0$ and
and 
$2EE\Trn E - \operatorname{trace}(EE\Trn)E = \Bz$.
Substituting in \Eqn{exyz} gives a system of homogeneous polynomial equations
which can be written as a 
$10\times20$ matrix ($M$) multiplied by the monomial vector,
$[x^3, y^3, x^2y, xy^2, x^2z, x^2, y^2z, y^2, xyz, xy, xz^{\!2},$ $ xz, x,
	yz^2, yz, y, z^3, z^2, z, 1]$.
\hilight{We have found that the most efficient way of computing the entries of 
$M$ is to use a computer algebra system to emit C code to build $M$
directly (see the
supplemental material).}

Gauss-Jordan reduction is applied to $M$, and a smaller matrix
$C(z)$ can be extracted which satisfies the homogeneous equations 
	$C(z) [x\ y\ 1]\Trn = \Bz$,
where the elements of $C(z)$ are degree 3 and 4 polynomials in $z$. Since $C$
has a null space, its determinant must be zero. Valid essential matrices for the
five matches are found by finding the roots of the degree 10 polynomial $d$
corresponding to $d(z) = |C(z)|$. 

Root finding is the single most expensive part of the algorithm.  As in
\cite{nister5point}, we use Sturm sequences to bracket the roots.  Following
the general philosophy of this paper, we tune our system to compute answers as
rapidly as possible even if it incurs the penalty of missing some valid
solutions.

\hilight{During bracketing, if a root is found to be at $|z|>100$, we abandon
any further attempts to find the root, since even if $z$ is found to machine
precision, $d$ will be far from zero. Additionally, we quickly abandon roots
which are quite close to being repeated, since such roots take a long time to
bracket and are numerically unstable.

For root polishing, we use the hybrid Newton-Raphson/Bisection algorithm given
in \cite{nr_in_C_root} for a maximum of 10 iterations, though it usually converges in
fewer than 6 iterations. If after 10 iterations or convergence, the value of $d$
is too far from zero, then the root is simply discarded.}

\begin{algorithm}
Pick a minimal set and find all valid  essential matrices using the algorithm
described above.
\label{alg:nister}
\end{algorithm}
\comment{This is the five point algorithm as described in~\cite{nister5point}
with some further speed optimizations. This technique yields zero to ten 
solutions.}

%

\section{Experiments and results}
\label{sec:results}

In this section results are given for all algorithms on a variety of synthetic
and real data.  The total computation required for the experiments was
approximately 500 CPU hours. For the robust algorithm (\Alg{robust}) we found that
10 point matches gave the best results. The results are computed in terms of
reliability, which is defined as the proportion of essential matrices found correctly.

\subsection{Synthetic data}

Synthetic frame pairs are generated for
a camera with a 90\degrees field of view, with translations up to 1
unit and rotations up to 35\degrees with the following method:

First, generate a point cloud so that points are distributed uniformly in the
first camera in position and inverse depth, starting depth of 1 unit.
Second, generate a random transform matrix and transform points to the second camera.
Then add Gaussian measurement noise ($\sigma=0.001$ units) to the projected position of the points in both cameras and
remove any points no longer visible.
Finally, generate a set of point matches from the points, create some mismatches (\ie outliers) and   
randomize the order of the points.
Regardless of the camera transformation, a
set number of good and bad matches are created.

From the data, we generate a fixed number of hypotheses and find the best one
using preemptive, breadth first RANSAC~\cite{nister03preemptive}. The best
hypothesis is then optimized on all the data using iterative reweighted least
squares with the Levenberg-Marquadt algorithm.  Unless specified, the results
are shown with the best preemptive RANSAC block
size~\cite{nister03preemptive}.

The total time is measured and averaged over 10,000 transformation matrices.
The final essential matrix is classified as correct or incorrect based on the
RMS (root mean square) reprojection error on the known inliers.  The
reliability is then
computed as the number of hypotheses is increased.

\begin{figure}
\includegraphics[width=0.495\textwidth]{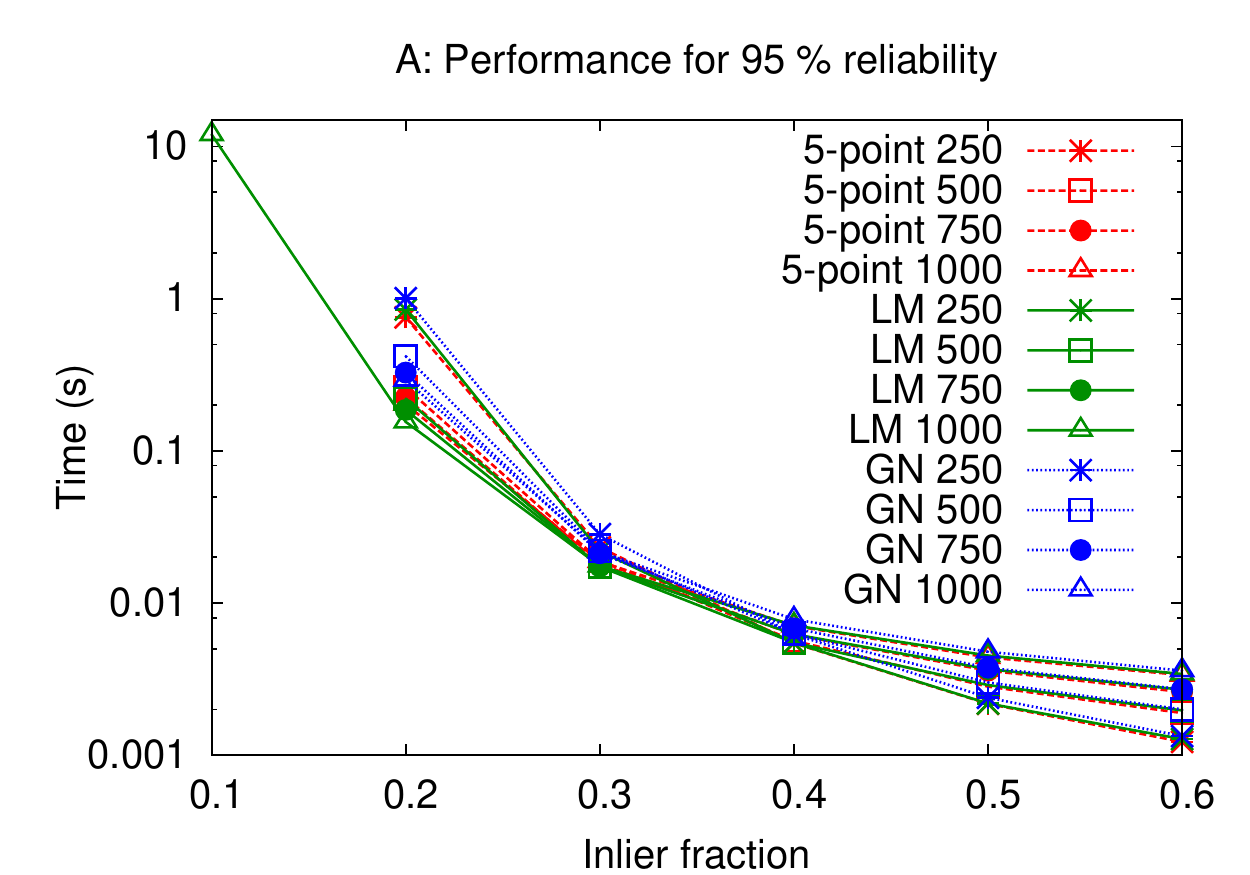}
\includegraphics[width=0.495\textwidth]{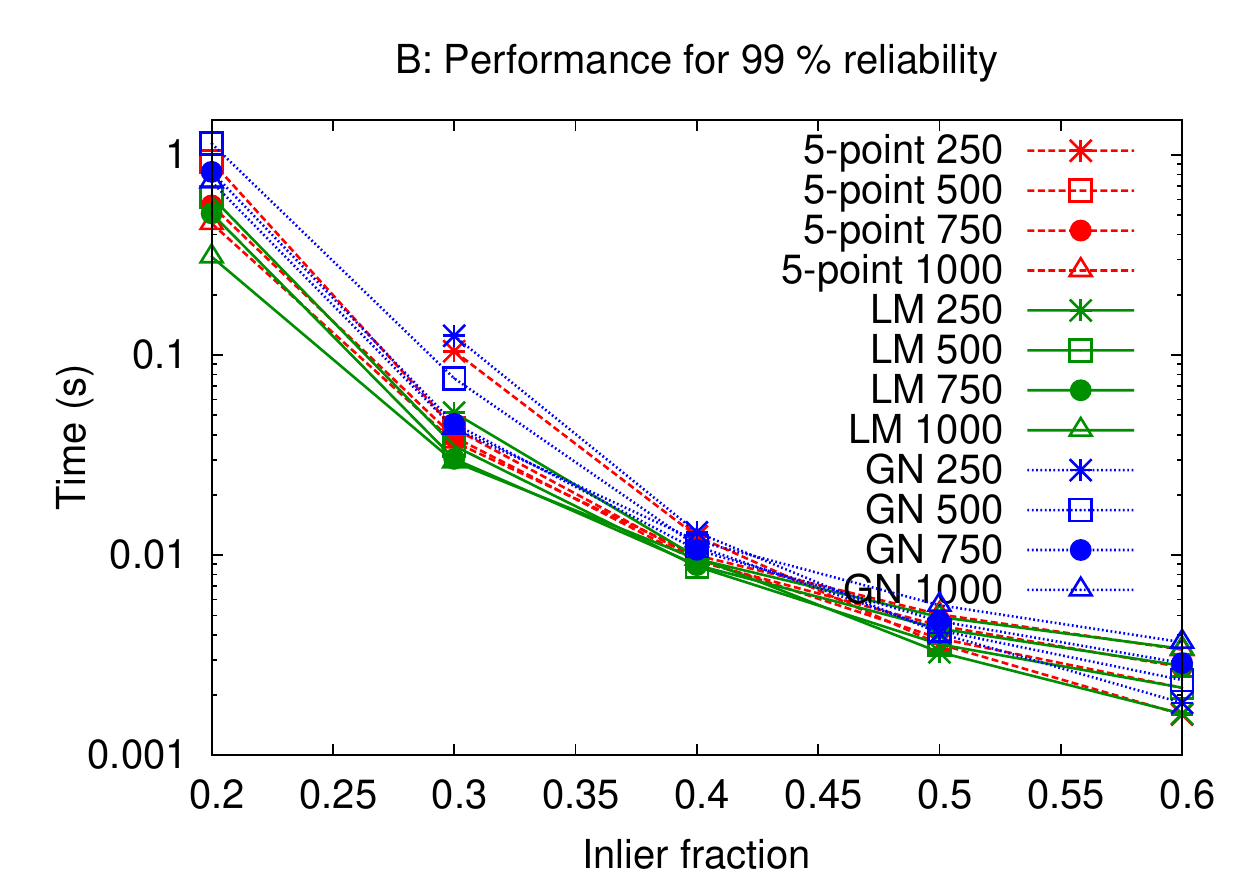}
\caption{
Graphs showing the time required to estimate an essential matrix with a given
reliability plotted for a given fraction of inliers. The plots are shown for
\Alg{lm} (LM), \Alg{nuts} (GN) and \Alg{nister} (5-point).  The number in the
legend denotes the total number of point matches.  Note that for 250 points per
frame, all of the algorithms require greater than 10 seconds to find a correct
essential matrix with 99\% reliability.  \label{fig:synres} } 
\end{figure}

The results are shown in \Fig{synres}, with the 
time required for a given reliability
plotted
against the 
inlier fraction. As can be seen \Alg{lm} is
the best performing with moderate proportion (up to 80\%) of outliers, outperforming
\Alg{nister} by about a factor of 1.5. In very low outlier situations, all
algorithms behave similarly, because other considerations (such as the final
optimization) start to dominate, though \Alg{nister} has a slight edge of about
2\% in some cases. However, even the very simple \Alg{nuts} performs very nearly
as well in these circumstances.

It is interesting to note that with low outlier densities it is better to have
few point matches but at high outlier densities, it is better to have more. As
one might expect, the optimal number of points decreases as the reliability
requirement is relaxed.

\begin{figure}
\includegraphics[width=0.495\textwidth]{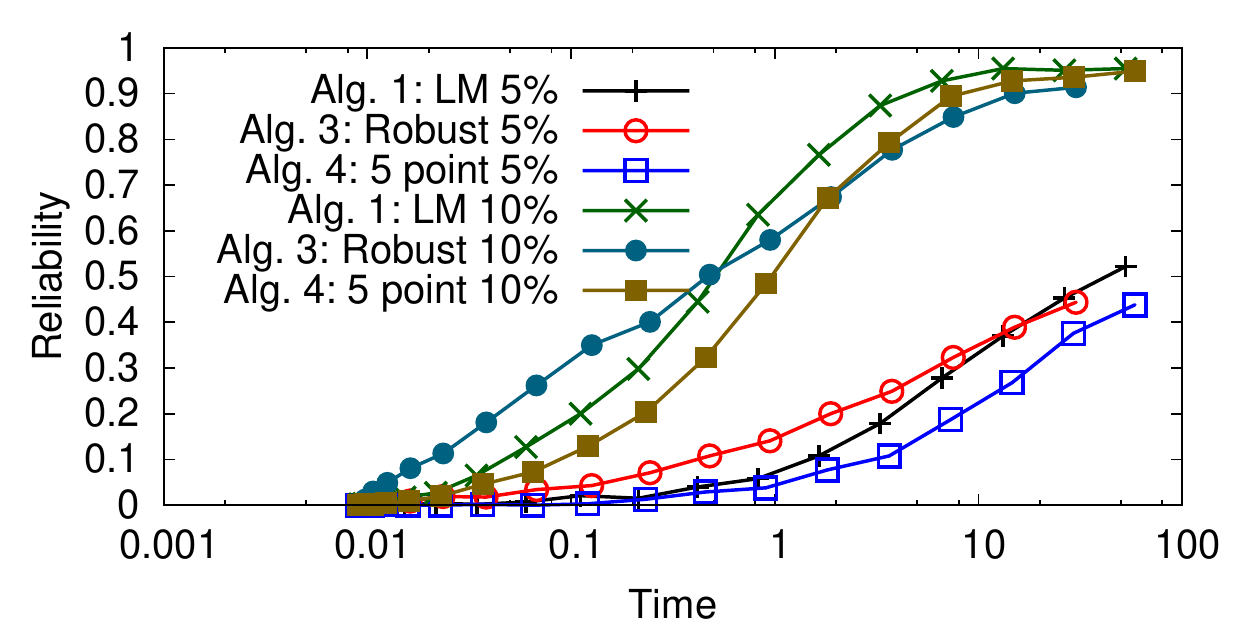}
\includegraphics[width=0.495\textwidth]{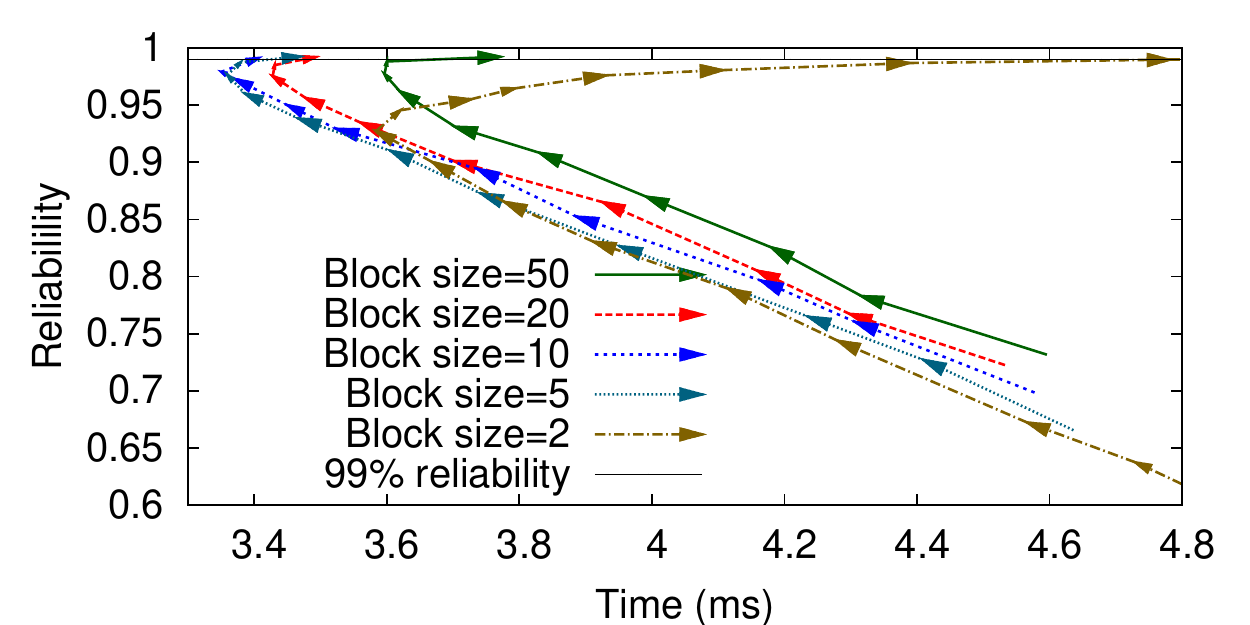}
\caption{
Plots of reliability against time for 1000 datapoints. Left: 10\% and 5\% inliers with 
a block size of 100. In this regime, \Alg{robust} is the best performing algorithm if
the computational budget is limited. Right: different block sizes, with 60\% inliers. Curve is parameterized
with the number of hypotheses.
In
this regime, the time spent in the nonlinear optimization at the end dominates.
Using a moderate number of hypotheses is faster overall than using a small
number of hypotheses since the time spent in the optimization is reduced.
\label{fig:hard}
\label{fig:synexample} 
} \end{figure}

\Alg{robust} is not shown in \Fig{synres} since it significantly underperforms the other
algorithms in this regime.  However, with a very high proportion of outliers, the
improved probability of picking a set of matches with at least 5 inliers
exceeds the relative slowness and low reliability of the algorithm, causing it
to dominate. This in shown in \Fig{hard}.

Another interesting point to note is that a reduction in the number of
hypotheses generated by RANSAC does not always reduce the processing time! A
striking example of this is shown in \Fig{synexample}. If a good starting point is
found, then the final robust optimization converges very quickly. However, if a
good starting point is not found, then the optimization can take a long time to
converge, and this computation dominates.  The effect is less pronounded in high
noise situations, eventually disappearing completely. 

In \Fig{synexample} A, all algorithms perform about equally well for 30\% inliers.
By comparison, \Alg{lm} evaluates about 800 hypotheses, \Alg{nuts} evaluates about 2,800 
and \Alg{nister} evaluates about 2,700 (and tests about 450 minimal sets).

%

\subsection{Real data}

\begin{figure*}
\def\Q{\hspace{.3em}}
\begin{tabular}{c@{\Q}c@{\Q}c@{\Q}c@{\Q}c@{\Q}c@{\Q}c@{\Q}c@{\Q}}
\includegraphics[width=0.11\textwidth]{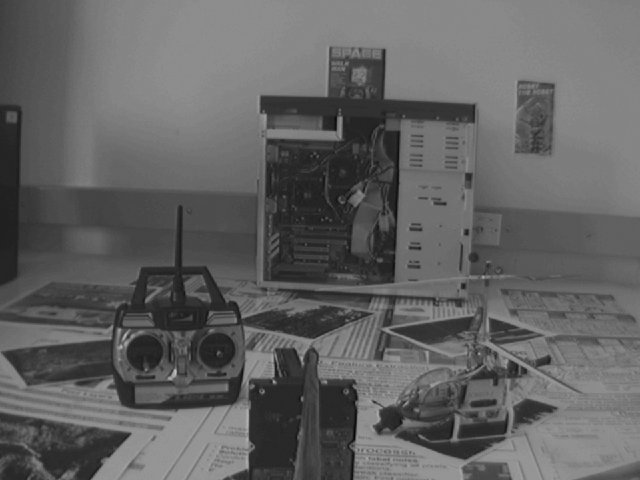}&
\includegraphics[width=0.11\textwidth]{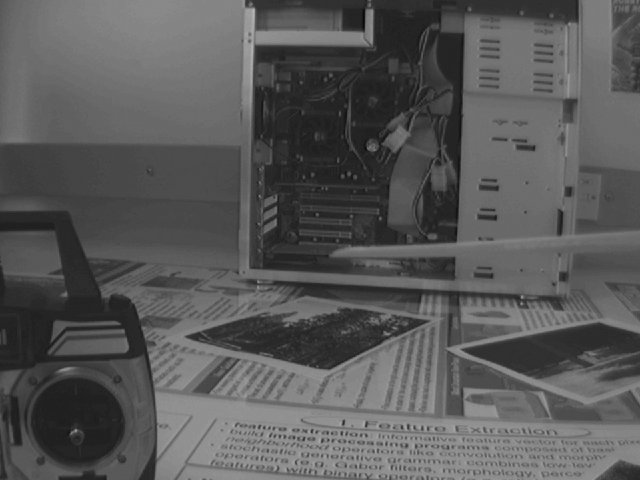}&
\includegraphics[width=0.11\textwidth]{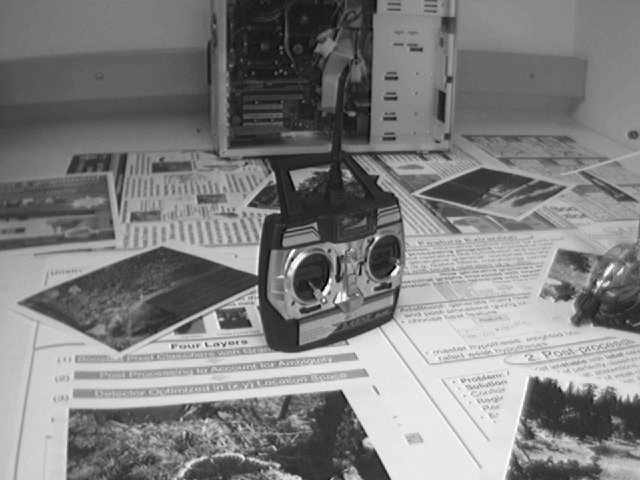}&
\includegraphics[width=0.11\textwidth]{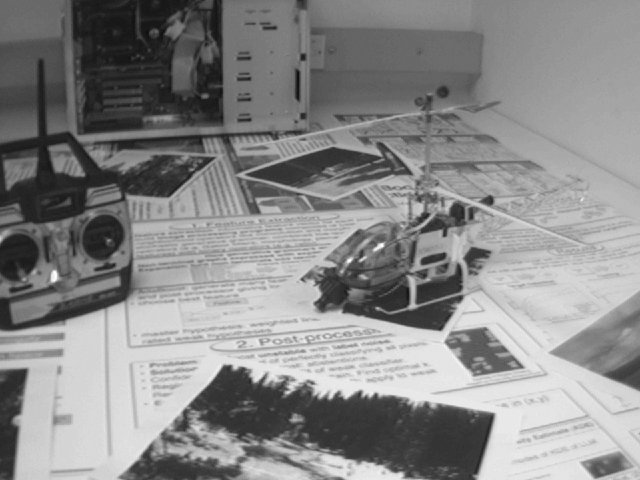}&
\includegraphics[width=0.11\textwidth]{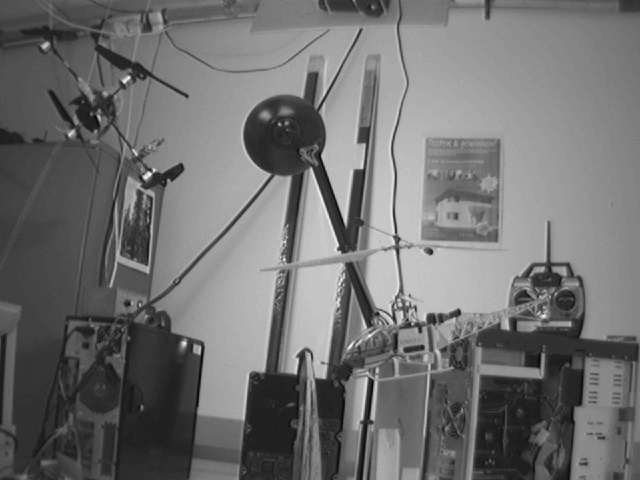}&
\includegraphics[width=0.11\textwidth]{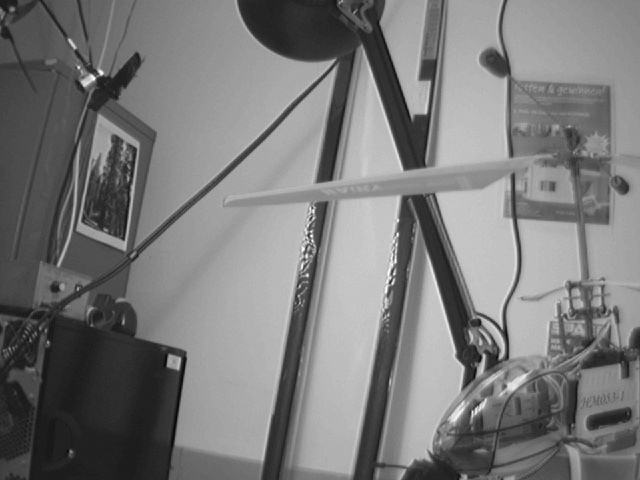}&
\includegraphics[width=0.11\textwidth]{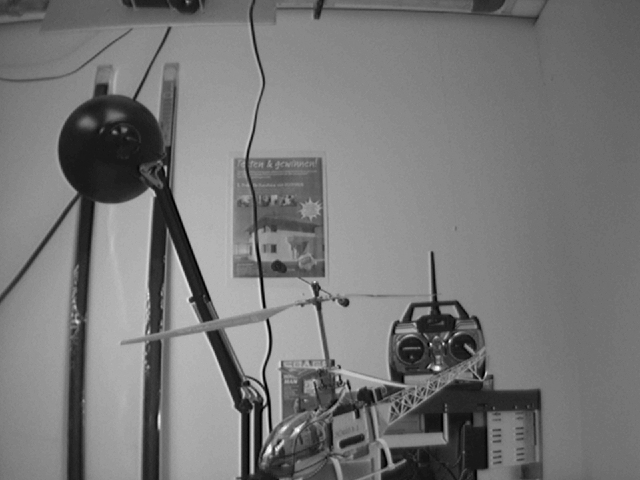}&
\includegraphics[width=0.11\textwidth]{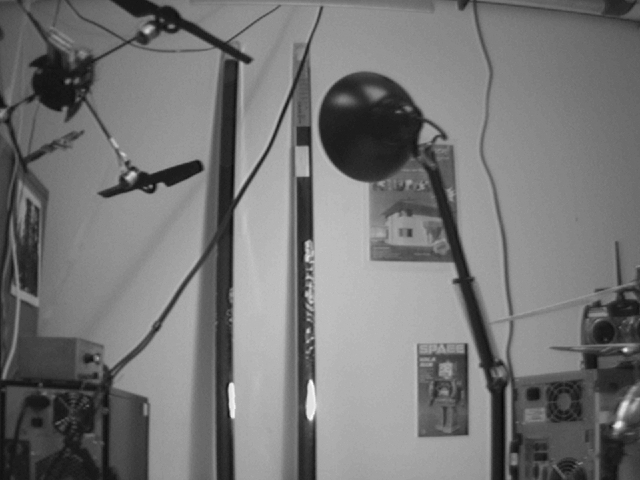}\\
A&B&C&D&E&F&G&H
\end{tabular}
\caption{A--D: dataset 1, dominant planar, textured structure. A, B optic
axis motion. C, D horizontal motion. E--H: dataset 2: no dominant plane or object
position. E, F optic axis motion. G, H horizontal motion.
\label{fig:dataset}}
\end{figure*}

\def\hd#1{\protect\rule[.53ex]{#1}{.5pt}}
\def\sk#1{\protect\rule[.53ex]{#1}{0pt}}
\def\ms#1#2{\hd{#1em}\sk{#2em}}
\def\sd{\ms{.3}{.3}}
\def\cd{\cdot\hspace{-2pt}}

\def\Key{
\def\Sss{\scriptsize}
\begin{tabular}{l@{\tiny}r@{\hspace{.1em}}c@{\hspace{.2em}}l@{\tiny}r@{\tiny}c@{\hspace{.2em}}l@{\tiny}r@{\tiny}c}
\Sss \Alg{lm}     &\Sss (LM)        & {\color{red}{\hd{1.0em}\protect\rule{1pt}{1.20ex}\hd{1.0em}}}& 
\Sss\Alg{nuts}   &\Sss (GN)        & {\color{NutsCol}\sd\sd\sd\hd{.3em}\Place{-1430,0}{$\times$}} &
\Sss\Alg{robust}\ \  &\Sss (Robust)    & {\color{RobuCol}{$\cd\cd\cd\cd\cd\cd\cd\hspace{-.3em}\raisebox{-.2ex}{$\Box$}\hspace{-.45em}\cd\cd\cd\cd\cd\cd$}}\\
\Sss\Alg{nister}\ \  &\Sss (5-point)   & {\color{blue}\sd\sd\sd\hd{.3em}\Place{-1230,0}{$\circ$}} &
\Sss Inlier ratio & (right axis)& {\color{InliCol}{\hd{2.40em}}}
\end{tabular}}

\setlength\unitlength{0.001em}
\def\Place#1#2{\protect\begin{picture}(0,0)\put(#1){#2}\end{picture}}
\definecolor{NutsCol}{rgb}{0.0,0.4,0.0}
\definecolor{RobuCol}{rgb}{0.75,0.0,0.75}
\definecolor{InliCol}{rgb}{0.00,0.56,0.56}
\begin{figure*}[f]
\includegraphics[width=0.33333\textwidth]{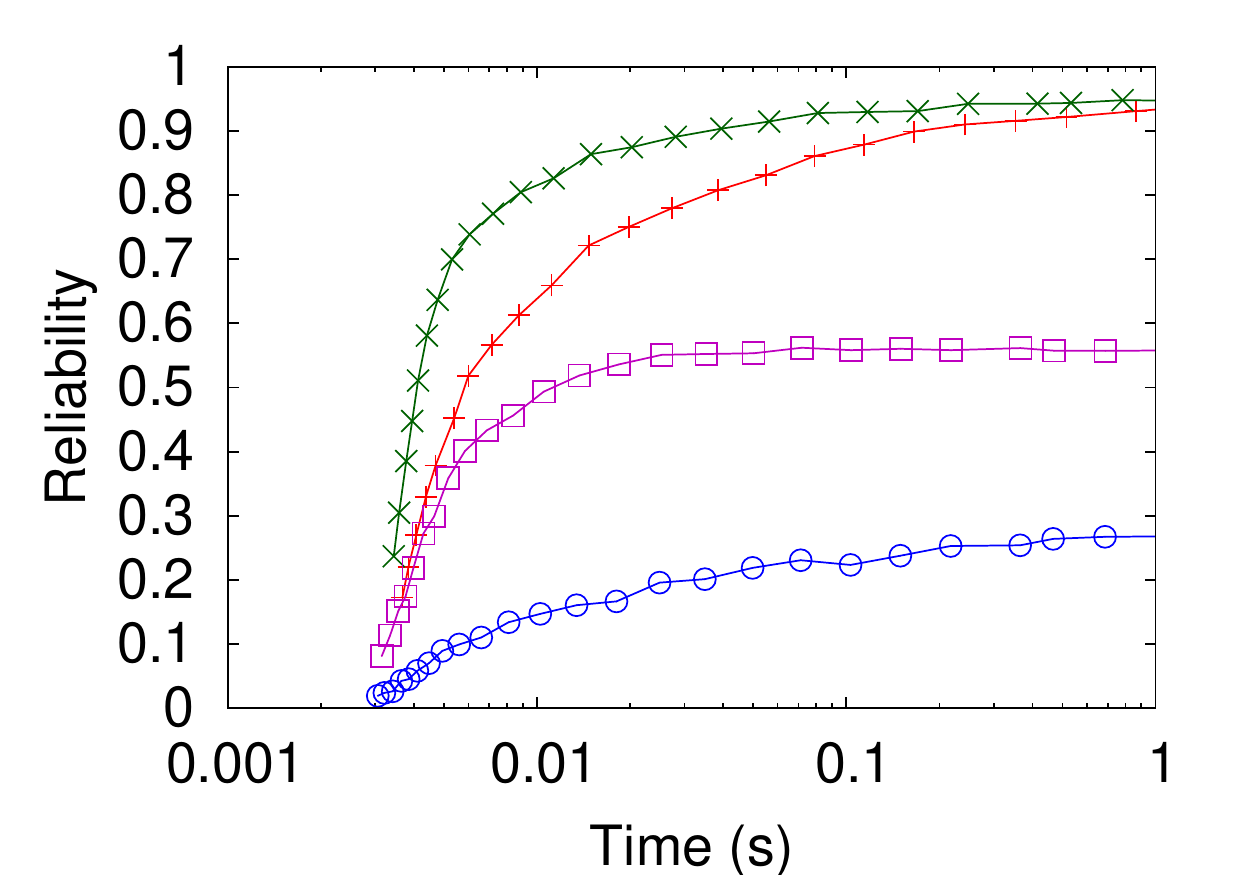}%
\includegraphics[width=0.33333\textwidth]{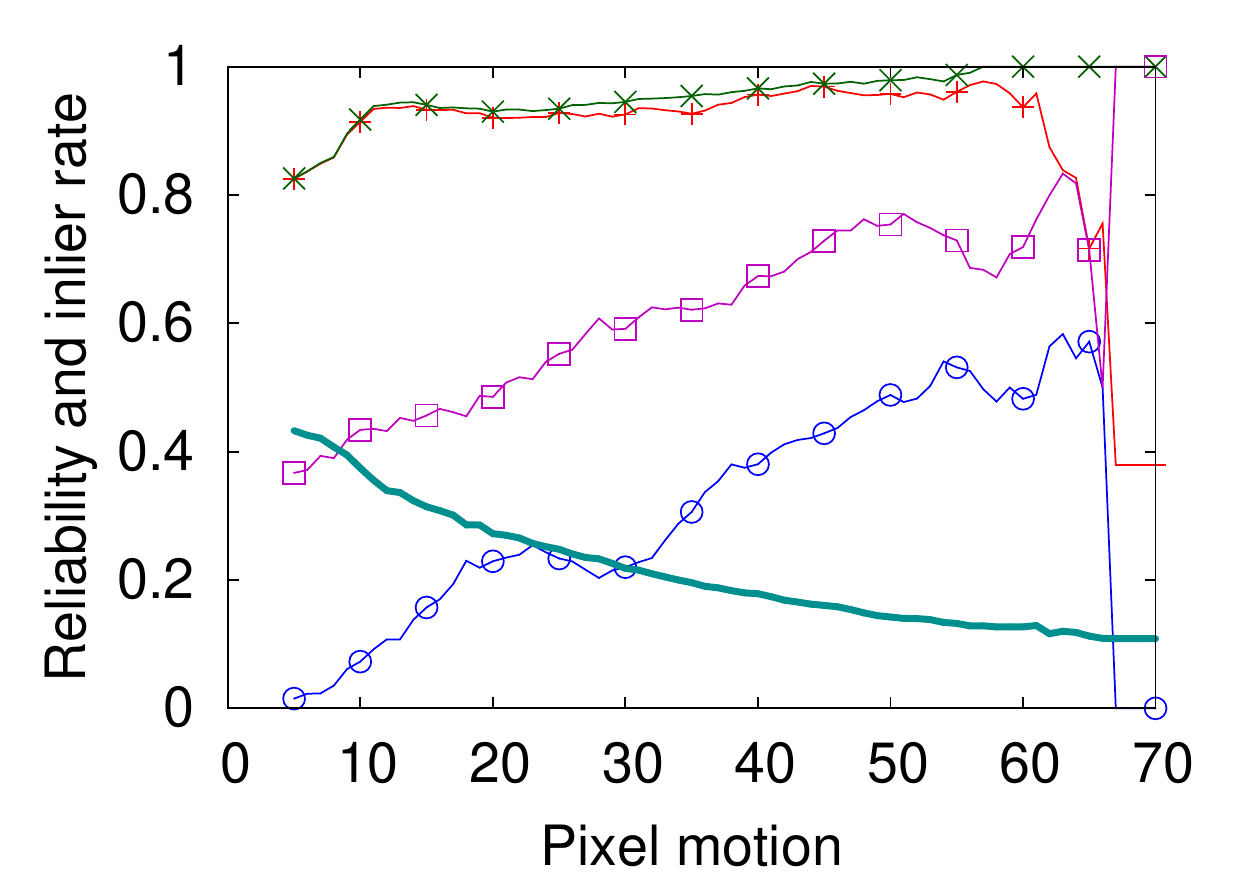}%
\includegraphics[width=0.33333\textwidth]{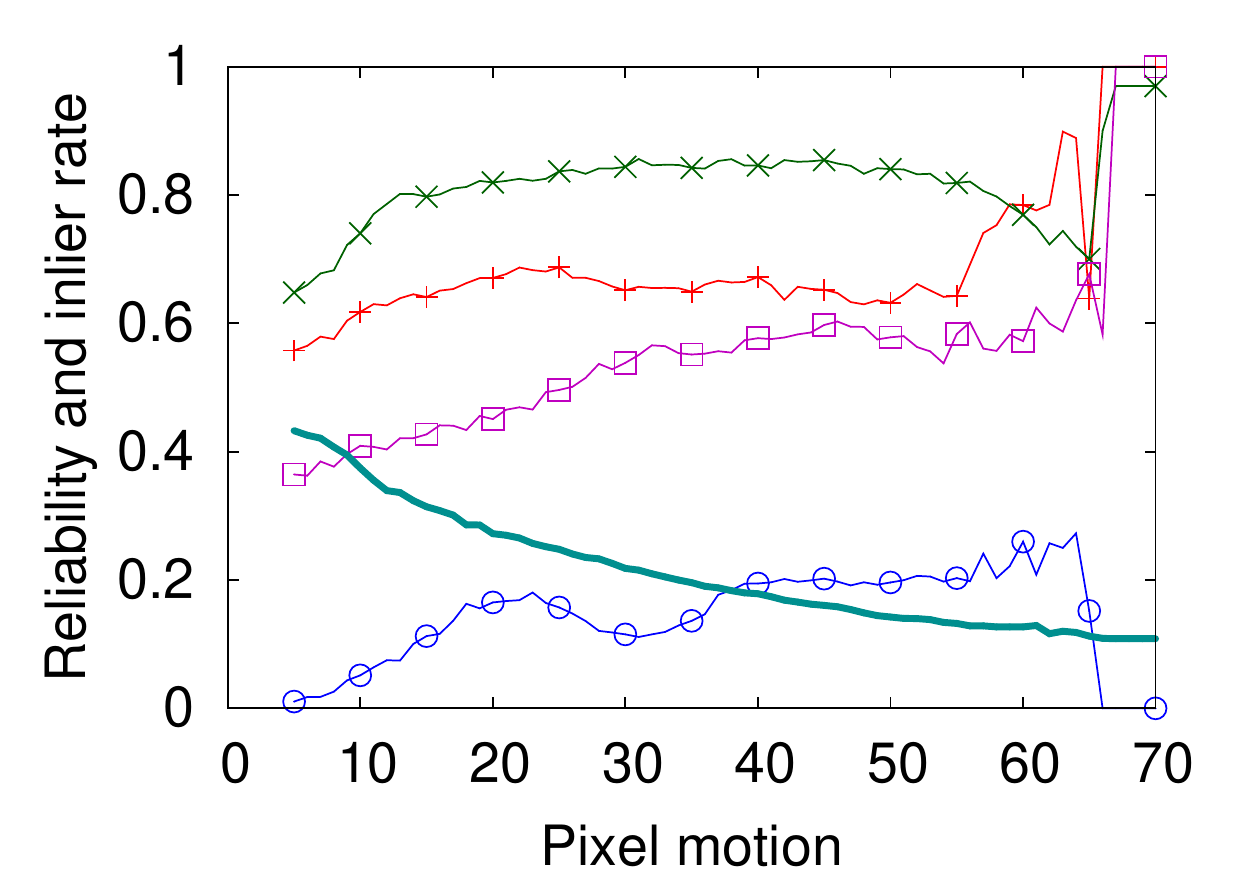}\\
\includegraphics[width=0.33333\textwidth]{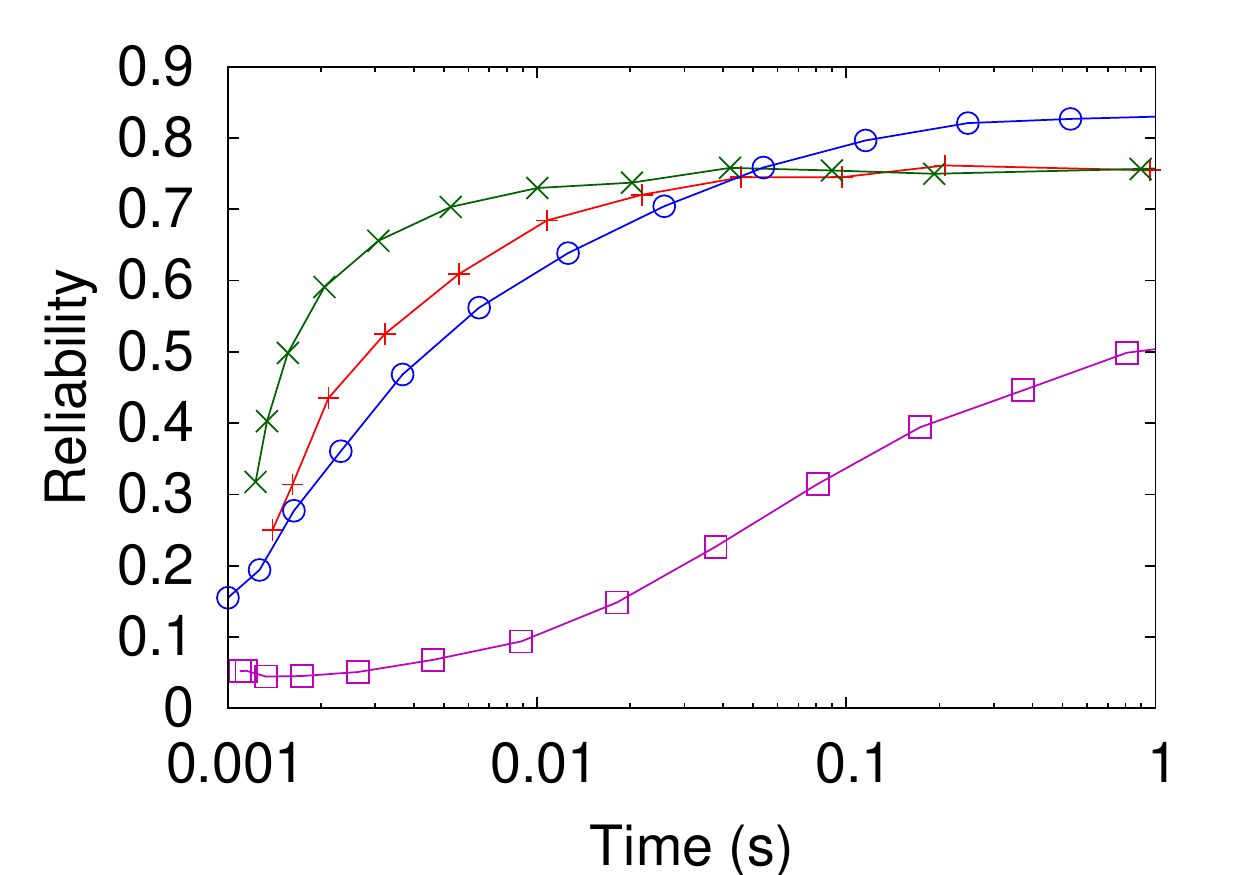}%
\includegraphics[width=0.33333\textwidth]{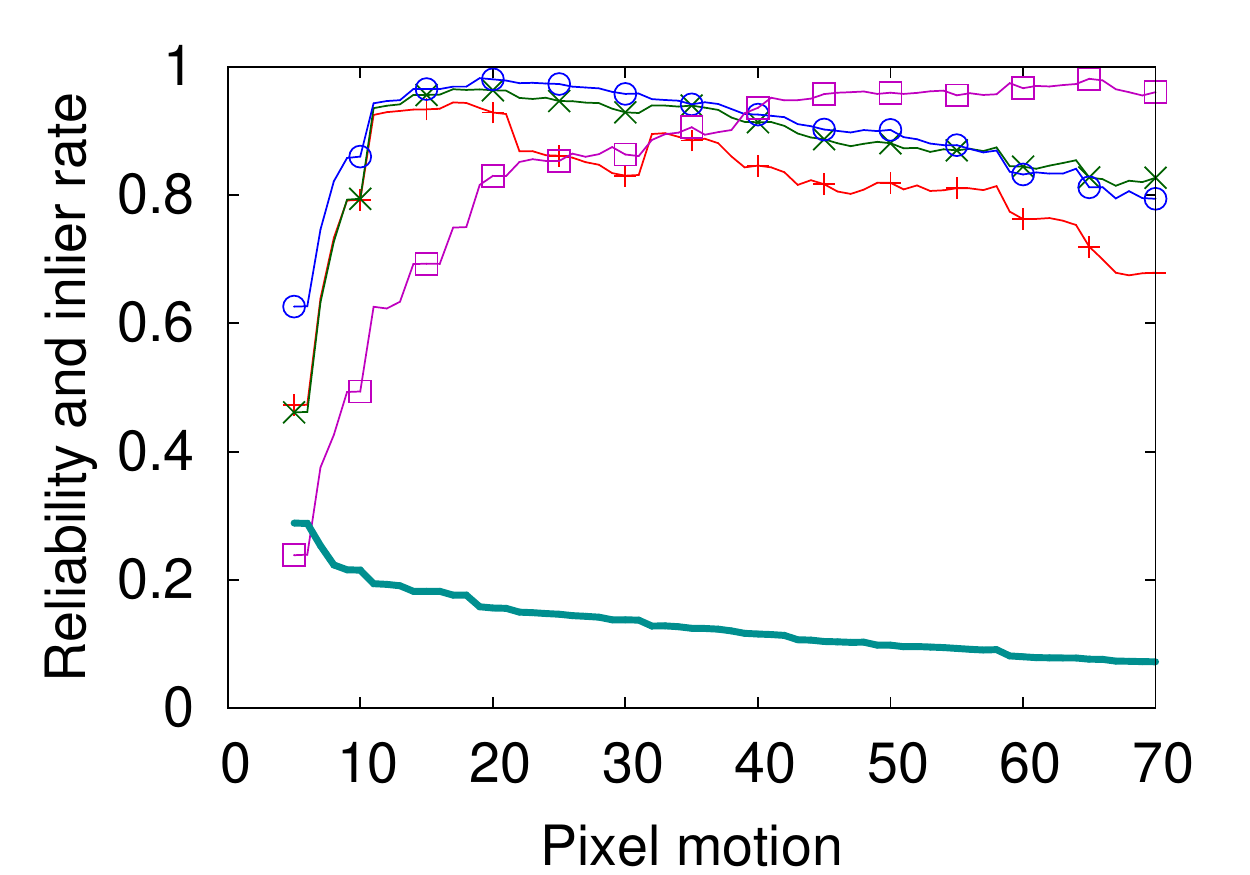}%
\includegraphics[width=0.33333\textwidth]{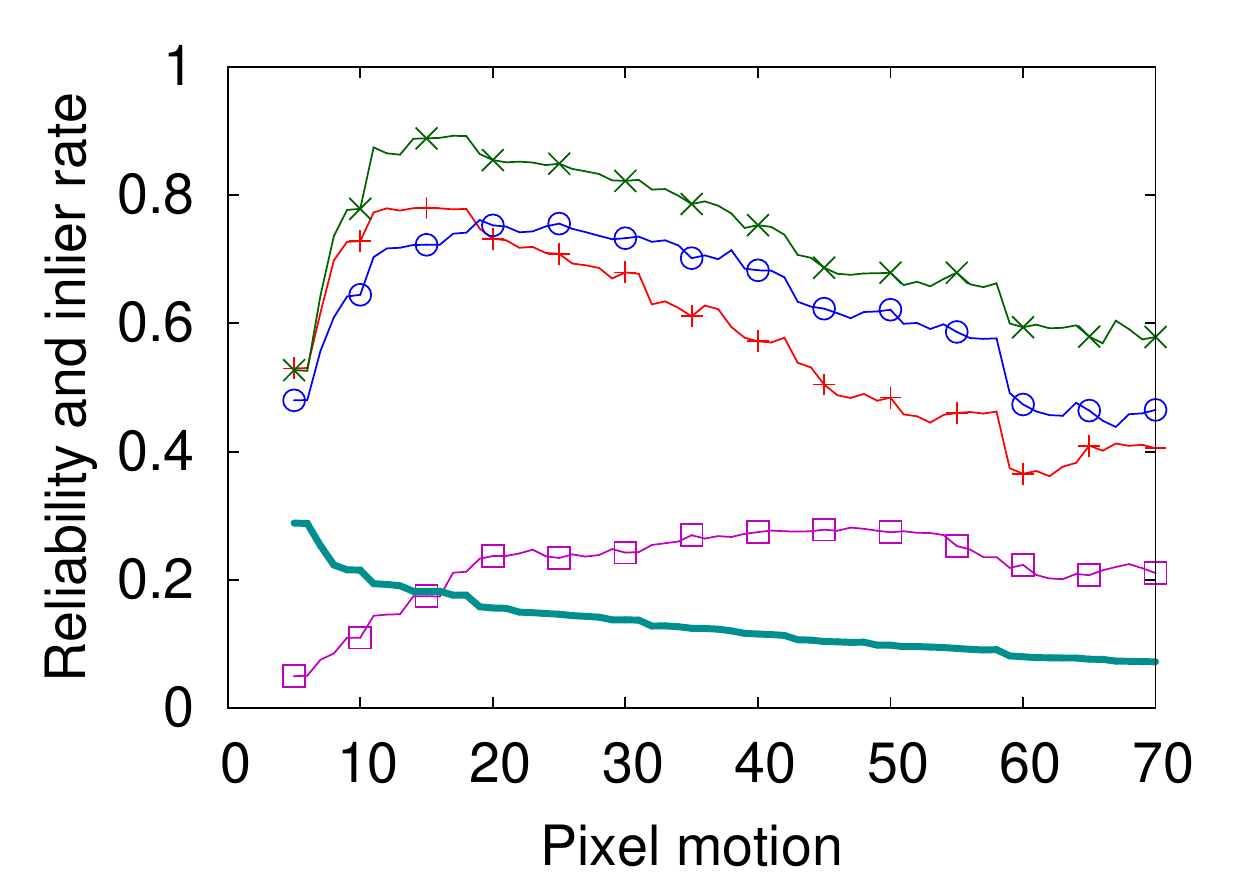}\\
\hrule
\includegraphics[width=0.33333\textwidth]{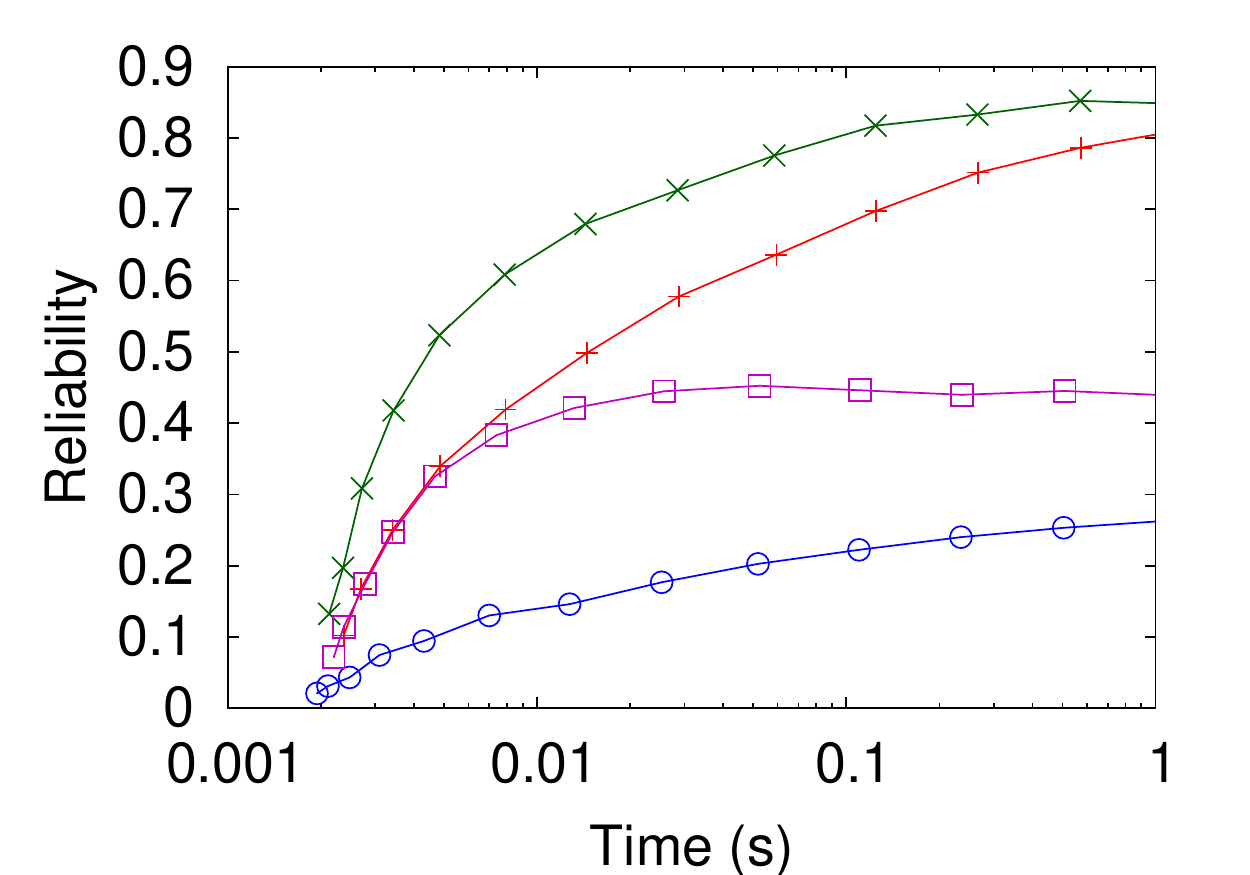}%
\includegraphics[width=0.33333\textwidth]{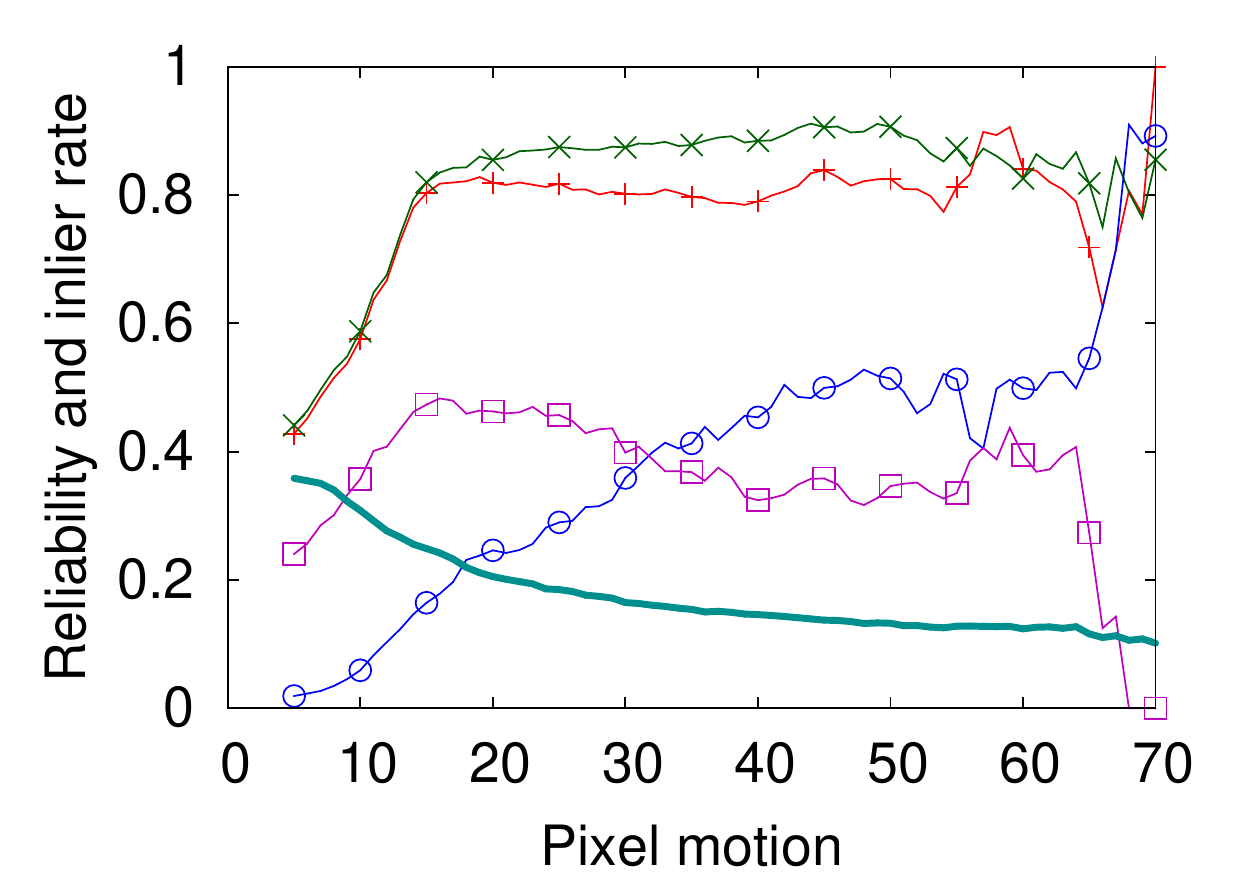}%
\includegraphics[width=0.33333\textwidth]{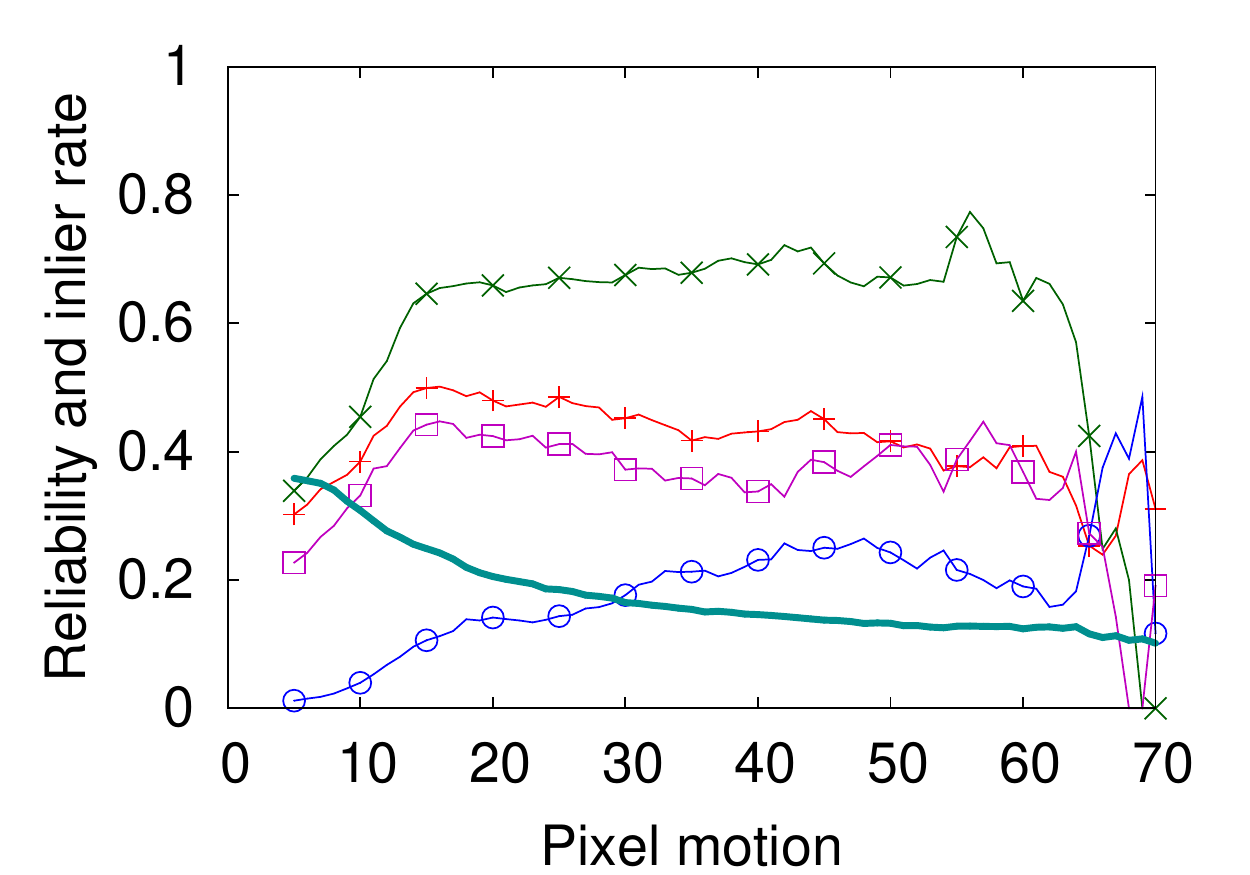}\\
\includegraphics[width=0.33333\textwidth]{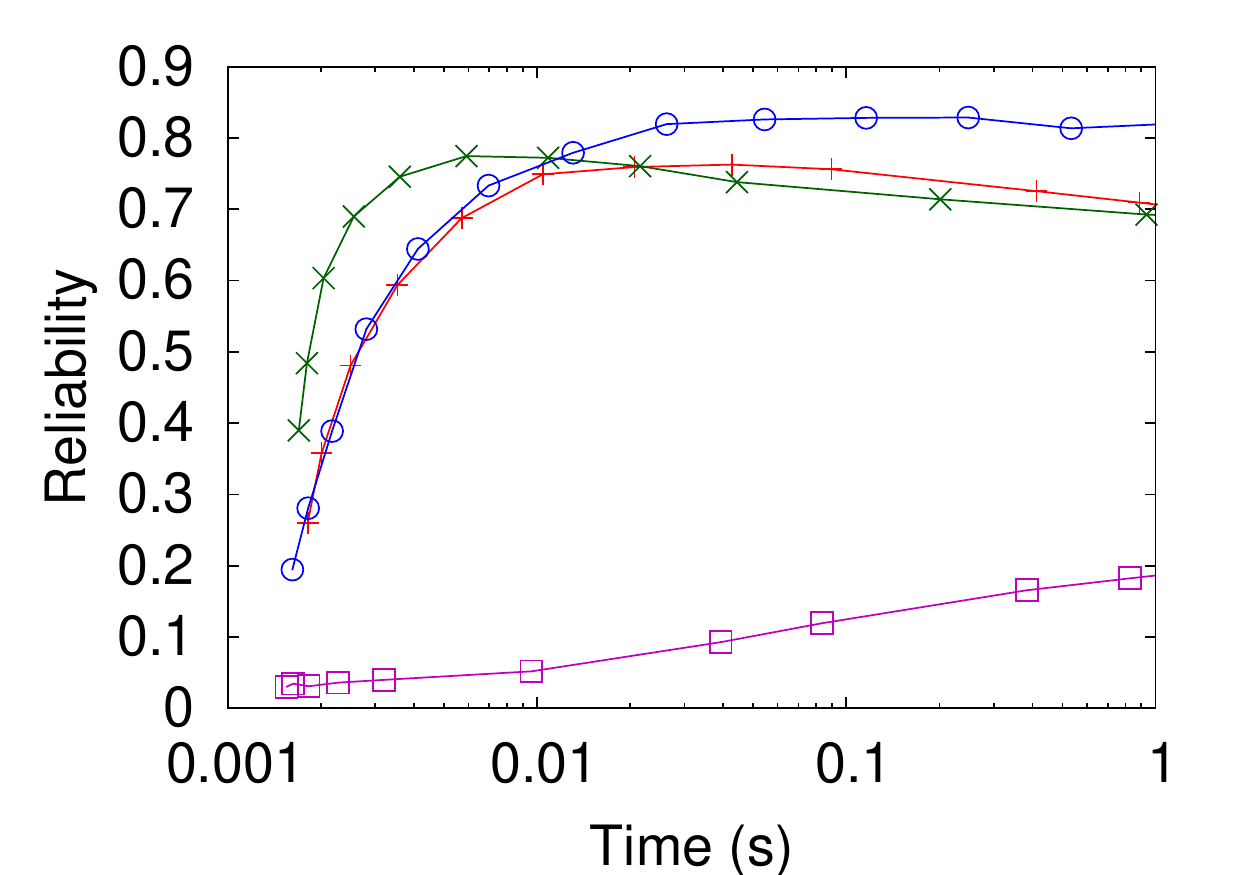}%
\includegraphics[width=0.33333\textwidth]{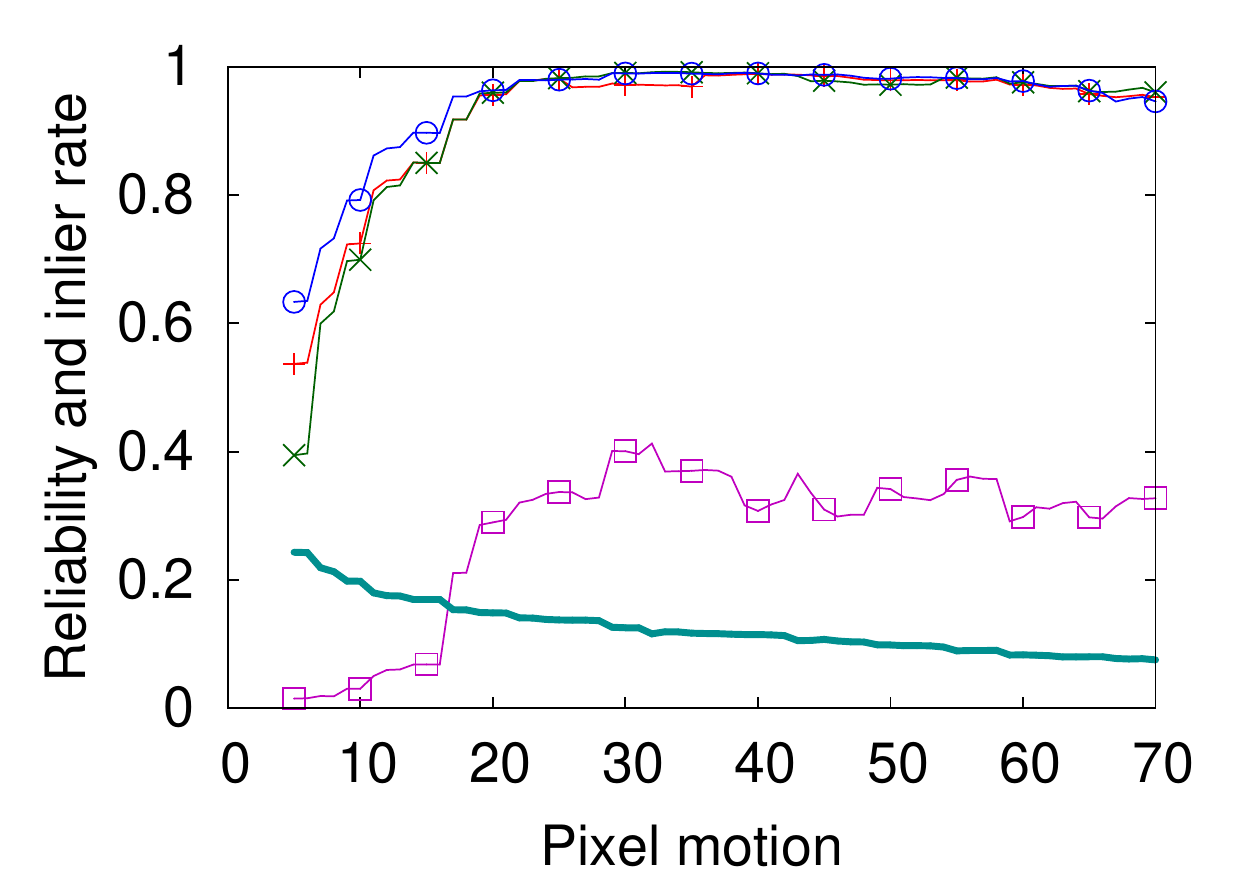}%
\includegraphics[width=0.33333\textwidth]{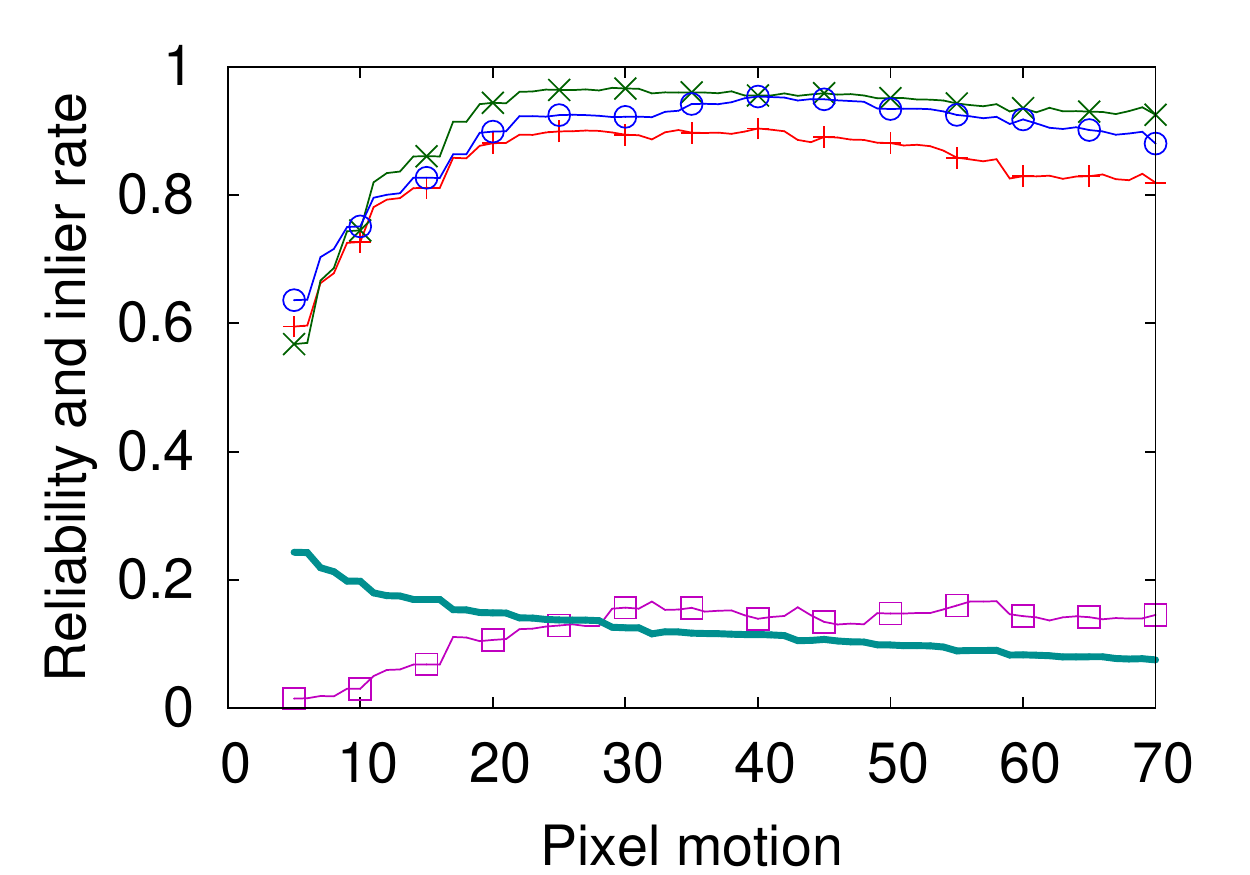}\\

\Key\\
\caption{Top half: results from dataset 1, $\sim500$ matches retained per
frame. Bottom half: results from dataset 2 ($\sim300$ matches).
Within each half: Top row:
horizontal motion. Bottom row: optic axis motion.  Left: reliability against
time aggregated over all data with at least 10 pixels of motion (inlier rate of
about 0.25).  Centre and Right: reliability plotted against average pixel
motion for 1 second per frame and 10ms per frame. Note that the inlier rate is
not constant, so it has been shown.
\label{fig:results}
} \end{figure*}

We generate real data by running a camera along a rail so that
the direction of motion is known.
Reconstructed essential matrices must be classified as
correct or incorrect, and this is done by thesholding on the angle between the
known translation direction and the reconstructed translation direction. We
choose 10\degrees as the threshold for the results shown, though the results
are similar for a range of thresholds. We do not threshold on rotation, since there
is some wobble in the motion of the camera.
The setup is arranged so that the
camera either has horizontal translation or translation along the optic axis.
Some sample images are shown in \Fig{dataset}. No {\it a priori} knowledge of this
is used in any of the algorithms.

The point correspondences are generated from a system which is designed
representative of a typical frame-rate vision application:
\begin{enumerate}
\item Generate an image pyramid with the scalings $\{1, \frac{2}{3},
\frac{1}{2}, \frac{2}{6}, \frac{1}{4}, \ldots\}$ since these ratios can be
generated very efficiently.
\item Perform FAST-9~\cite{fast-paper} on each layer of the pyramid and extract a
feature for each corner.
\item For each corner in the frame at time $t$, find the best match in the frame
at time $t-N$, with no restriction on matching distance.
\item The best 20\% of matches are retained and their order is randomized.
\end{enumerate} 
We take $N \in \{1..60\}$ to increase the baseline, making the number of frame
pairs tested about 15,000. Note that the value of $N$ is not used in the 
creation of any of the results below.
The essential matrix is then
estimated using RANSAC followed by an M-estimation step. We extract the
translation and rotation using Horn's method~\cite{horn90recovering} and
triangulate points to determine which of the four combinations of rotation and
translation to use.

Results for dataset 1 are shown in \Fig{results}.  As can be seen, unlike in
the synthetic data, the simplest algorithm employing a Gauss-Newton optimizer
exhibits exceptional performance compared to the other algorithms. The
performance increase relative to the LM optimizer is because the GN optimizer
very quickly abandons sets of points which are hard to optimize. As a result,
it tests many more minimal sets than the LM optimizer.

With the optic axis motion, the five-point algorithm performs much better.
This motion is problematic for
optimization based techniques since they are prone to local minima because the
movement of the epipoles causes dramatic motion of the epipolar lines. A direct
method such as the five-point algorithm does not suffer from this effect.

The results show that the estimation of the essential matrix is particularly
difficult when the optic flow is small. As can be seen, the reliability
decreases slowy (or even increases) with increasing optic flow, even though the
inlier rate drops significantly. The drop in inlier rate would cause a very
large drop in performance if the reliability of the algorithms did not increase
dramatically with inlier rate.

As the pixel motion gets large, the inlier rate drops since feature point
matching becomes more difficult. In these regimes, \Alg{robust} (robust
estimation) shows some significant improvements over the other algorithms.

The experiments on dataset 1 were repeated with higher corner detection
thresholds giving 300, 200 and 100 retained matches per frame. The performance
generally decreased with increasing thresholds, but the trends were largely
unaffeced.

Results for dataset 2 are 
quite similar to dataset 1. The main difference is that \Alg{nister}
performs somewhat better relative to dataset 1 and \Alg{robust} performs
somewhat worse.

Finally, we repeated the experiments with a less accurate camera calibration.
The inaccuracy caused a slight performance decrease across all algorithms. No
algorithm appeared to be significantly less stable than any other in the presence
of small calibration errors.

\section{Conclusions}

In general, reliable estimation of essential matrices remains very difficult
problem. {\bfseries\textsl{On real data, the simplest algorithm---generating
RANSAC hypotheses by minimizing residuals of a minimal set with Gauss-Newton
(\Alg{nuts})---outperforms the other algorithm by a wide margin.}}

On the synthetic data, LM (\Alg{lm}), GN (\Alg{nuts}) and the five point
algorithm (\Alg{nister}) perform similarly, with \Alg{lm} winning by a
relatively wide margin with high outlier densities and \Alg{nister} winning by a
small margin at low outlier densities. We also note that as expected, the
performance of the robust algorithm (\Alg{robust}) is best when the outlier
density is very high, proving to be the most suitable algorithm in high
noise-time constrained situations.

The results on real data are somewhat different and serve as a good illustration
as to the pitfalls of relying too heavily on synthetic data. The main point is
that \Alg{nuts} is by far the best performer when it comes to reconstructing
left-right motions. This is particularly interesting given that is also by far
the simplest algorithm to implement. The case is less clear cut for
forward-backward motions, with \Alg{nister} winning by a considerable margin in
some cases. Additionally, \Alg{robust} can perform better than all other
algorithms in high noise situations.

In conclusion, for essential matrix estimation, simple optimization with
Ga\-uss-Newton (\Alg{nuts}) is the best
performing algorithm, giving the most consistently reliable results, especially
in time-constrained operation. If computation time is not at a premium, then the
best results would probably be achieved by a system which draws hypotheses from
\Alg{nuts} and \Alg{nister}.
These results also have wider applicability: simple, fast and numerically stable
iterative algorithms can be used for generating hypotheses for RANSAC
in many
situations, including those where currently complex, direct solutions are used
and those for which no direct solutions are known.



\clearpage
\bibliography{papers}
\end{document}